\documentclass[sigconf,nonacm]{acmart}
\AtBeginDocument{%
  }

\setcopyright{acmlicensed}
\copyrightyear{2018}
\acmYear{2018}
\acmDOI{XXXXXXX.XXXXXXX}
\acmConference[Conference acronym 'XX]{Make sure to enter the correct
  conference title from your rights confirmation email}{June 03--05,
  2018}{Woodstock, NY}
\acmISBN{978-1-4503-XXXX-X/2018/06}

\usepackage{graphicx}  % 用于识别 \includegraphics 插入图片
\usepackage{booktabs}  % 用于绘制顶会标准的“三线表” (\toprule, \midrule, \bottomrule)
\usepackage{multirow}  % 核心！用于支持跨行合并 (\multirow{行数}{*}{内容})
\usepackage{amsmath}   % 用于支持高级数学符号和排版
\acmSubmissionID{1712}

\begin{document}

\title{UniEditBench: A Unified and Cost-Effective Benchmark for Image and Video Editing via Distilled MLLMs}

\title{UniEditBench: A Unified and Cost-Effective Benchmark for Image and Video Editing via Distilled MLLMs}

%% 作者列表开始
\author{Lifan Jiang}
\affiliation{%
  \institution{Zhejiang University}
  \city{Hangzhou}
  \country{China}}
\email{lifanjiang@zju.edu.cn}

\author{Tianrun Wu}
\affiliation{%
  \institution{Zhejiang University}
  \city{Hangzhou}
  \country{China}}
\email{tianrunwu@zju.edu.cn}

\author{Yuhang Pei}
\affiliation{%
  \institution{Zhejiang University}
  \city{Hangzhou}
  \country{China}}
\email{3200104179@zju.edu.cn}

\author{Chenyang Wang}
\affiliation{%
  \institution{Zhejiang University}
  \city{Hangzhou}
  \country{China}}
\email{yongyuan_chen@zju.edu.cn}

\author{Boxi Wu}
\affiliation{%
  \institution{Zhejiang University}
  \city{Hangzhou}
  \country{China}}
\email{wuboxi@zju.edu.cn}

\author{Deng Cai}
\affiliation{%
  \institution{Zhejiang University}
  \city{Hangzhou}
  \country{China}}
\email{dengcai@gmail.com}

%% 如果需要设置共同贡献或通讯作者，可以使用以下命令：
%% \authornote{Both authors contributed equally to this research.}
%% \authornotemark[1]

\renewcommand{\shortauthors}{Jiang et al.}


\begin{abstract}
The evaluation of visual editing models remains fragmented across methods and modalities. Existing benchmarks are often tailored to specific paradigms, making fair cross-paradigm comparisons difficult, while video editing lacks reliable evaluation benchmarks. Furthermore, common automatic metrics often misalign with human preference, yet directly deploying large multimodal models (MLLMs) as evaluators incurs prohibitive computational and financial costs. We present \textbf{UniEditBench}, a unified benchmark for image and video editing that supports reconstruction-based and instruction-driven methods under a shared protocol. UniEditBench includes a structured taxonomy of nine image operations (Add, Remove, Replace, Change, Stroke-based, Extract, Adjust, Count, Reorder) and eight video operations, with coverage of challenging compositional tasks such as counting and spatial reordering. To enable scalable evaluation, we distill a high-capacity MLLM judge (Qwen3-VL-235B-A22B Instruct) into lightweight 4B/8B evaluators that provide multi-dimensional scoring over structural fidelity, text alignment, background consistency, naturalness, and temporal-spatial consistency (for videos). Experiments show that the distilled evaluators maintain strong agreement with human judgments and substantially reduce deployment cost relative to the teacher model. UniEditBench provides a practical and reproducible protocol for benchmarking modern visual editing methods. Our benchmark and the associated reward models are publicly available at \url{https://github.com/wesar1/UniEditBench}.
\end{abstract}


\begin{CCSXML}
<ccs2012>
   <concept>
       <concept_id>10010147.10010178</concept_id>
       <concept_desc>Computing methodologies~Artificial intelligence</concept_desc>
       <concept_significance>500</concept_significance>
       </concept>
   <concept>
       <concept_id>10010147.10010178.10010224.10010240.10010241</concept_id>
       <concept_desc>Computing methodologies~Image representations</concept_desc>
       <concept_significance>300</concept_significance>
       </concept>
 </ccs2012>
\end{CCSXML}

\ccsdesc[500]{Computing methodologies~Artificial intelligence}
\ccsdesc[300]{Computing methodologies~Image representations}

\keywords{Visual Editing, Benchmark, Distillation, Cost-Effective Evaluation}
\begin{teaserfigure}
\centering
  \includegraphics[width=0.96\textwidth]{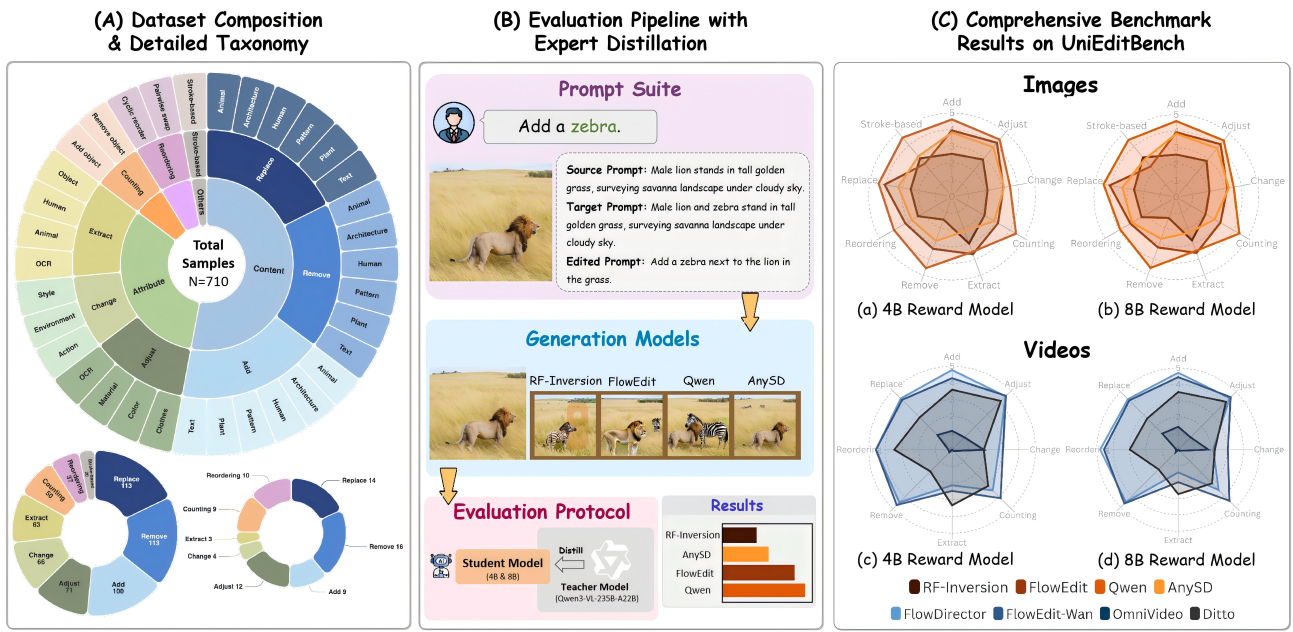}
  \vspace{-0.1em}
  \caption{Overview of the UniEditBench framework. (A) Dataset Composition \& Detailed Taxonomy: Displays the structured task taxonomy and data distribution, comprising 633 image and 77 video samples across diverse editing operations (e.g., content modification, attribute transformation, and compositional reasoning tasks like counting and reordering). (B) Evaluation Pipeline with Expert Distillation: Our pipeline provides a source prompt, target prompt, and edited prompt for each sample pair to accommodate various editing paradigms. Additionally, the editing results, original images, and prompt information are input into our 4B/8B discriminative models, which are distilled from the Qwen3-VL-235B-A22B model, to obtain the final judgments. (C) Comprehensive Benchmark Results: Radar charts display the aggregated evaluation results of editing methods on UniEditBench, assessed by the distilled 4B and 8B reward models (top: image benchmark; bottom: video benchmark).}
  \Description{Enjoying the baseball game from the third-base
  seats. Ichiro Suzuki preparing to bat.}
  \label{fig:teaser}
  \vspace{-0.1em}
\end{teaserfigure}


\settopmatter{printacmref=false}
\renewcommand\footnotetextcopyrightpermission[1]{}
\maketitle

\section{Introduction}
\vspace{-1pt}

Recent advances in visual generative models---especially diffusion-based frameworks~\cite{ho_ddpm, rombach_stable_diffusion} and multi-modal large language models (MLLMs)~\cite{liu_llava, qwen_vl_series}---have substantially improved visual content creation. Beyond image synthesis, current research emphasizes user-controllable and fine-grained editing, including object replacement, attribute manipulation, and motion-aware video editing~\cite{hertz_prompt2prompt, brooks_instructpix2pix, zhang_controlnet}. As a result, visual editing is becoming an increasingly important component of interactive multimedia systems.

Despite this rapid progress, evaluation remains fragmented and often incomparable across method families. We observe three practical gaps. \textbf{(1) Paradigm fragmentation.} Existing benchmarks are usually optimized for a subset of models, such as reconstruction-based methods (inversion-based~\cite{diffedit, rf_inversion} and inversion-free~\cite{flowedit, snr_edit}) or instruction-driven methods (structured editing and MLLM-based approaches~\cite{brooks_instructpix2pix, qwen_image}), which limits fair cross-paradigm comparison. \textbf{(2) Limited video coverage.} Relative to image editing, benchmark design and protocol standardization for video editing are still limited, although temporal coherence and inter-frame consistency are central to real-world quality~\cite{sora, videocrafter}. \textbf{(3) Evaluator cost--fidelity tension.} Conventional automatic metrics (e.g., PSNR, CLIP-based scores~\cite{radford_clip}) are efficient but may not fully reflect human preference, while directly using very large MLLMs as evaluators can impose substantial computational and financial cost.

\begin{figure*}[!t]
    \centering
    \includegraphics[width=0.96\textwidth]{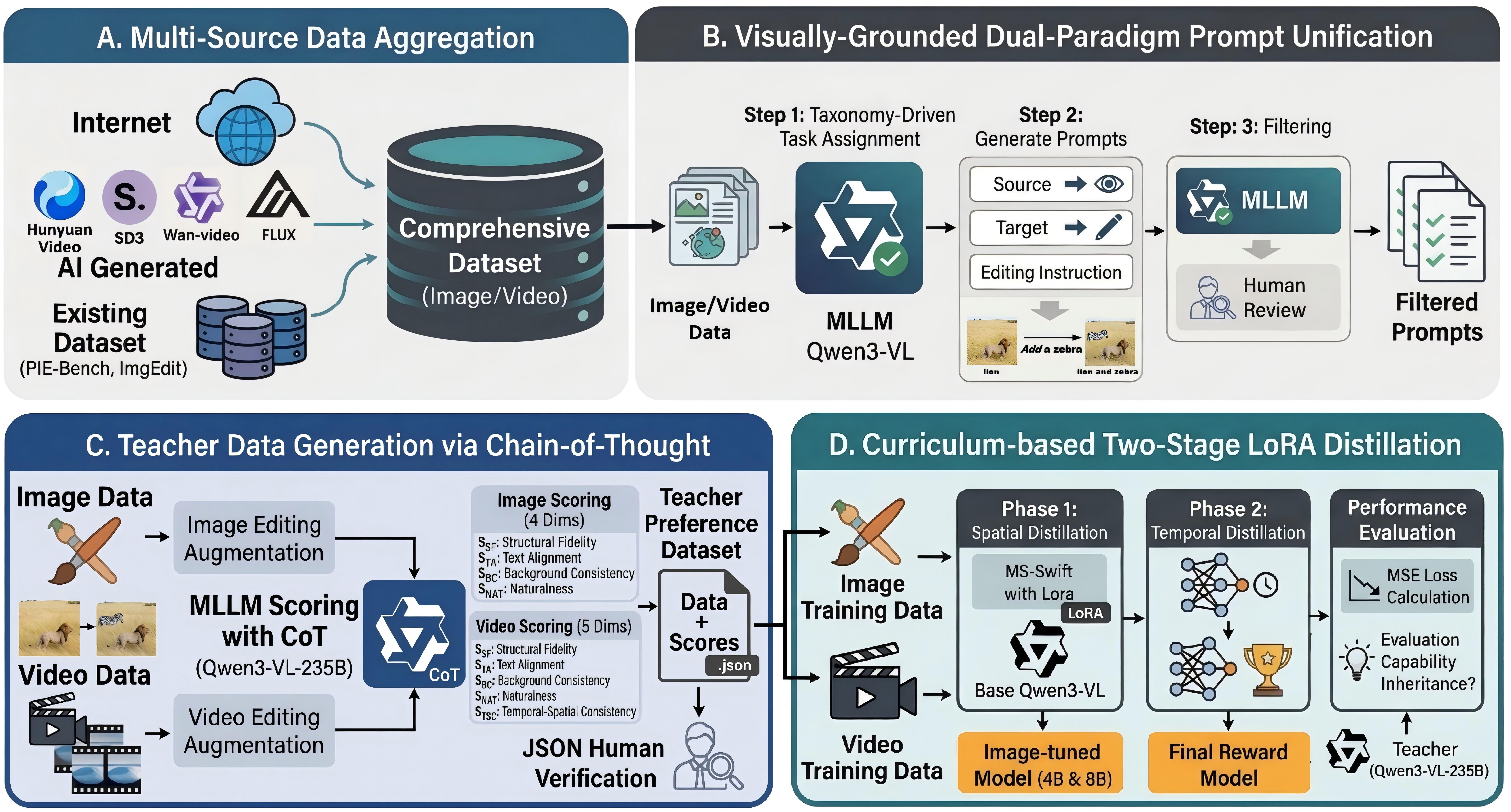}
    \caption{The overall pipeline of UniEditBench. (A) Multi-Source Data Aggregation: A comprehensive dataset is constructed by aggregating assets from the internet, AI generation models (e.g., FLUX, SD3, Wan-video\cite{wan2025wan}, HunyuanVideo\cite{hunyuanvideo}), and existing benchmarks (PIE-Bench, ImgEdit). (B) Visually-Grounded Dual-Paradigm Prompt Unification: Qwen3-VL is employed to assign taxonomy tasks and generate prompt triplets (source, target, editing instruction) to bridge reconstruction and instruction-driven paradigms, followed by strict MLLM and human filtering. (C) Teacher Data Generation via Chain-of-Thought: Image and video samples undergo editing augmentation and are scored by a teacher model (Qwen3-VL-235B) using CoT reasoning across multiple dimensions to form a preference dataset. (D) Curriculum-based Two-Stage LoRA Distillation: Student models (4B \& 8B) are fine-tuned via the MS-Swift framework using a two-phase curriculum: spatial distillation on images followed by temporal distillation on videos, preventing catastrophic forgetting while yielding cost-effective evaluator models.}
    \label{fig:pipeline}
\end{figure*}

To address these issues, we present \textbf{UniEditBench}, a unified benchmark for evaluating both image and video editing models under a shared protocol. UniEditBench supports reconstruction-based and instruction-driven paradigms through a common taxonomy of nine image operations (Add, Remove, Replace, Change, Stroke-based, Extract, Adjust, Count, Reorder) and eight video operations. The benchmark is designed not only for appearance-level edits, but also for reasoning-intensive settings such as counting and spatial reordering, which are frequently under-represented in prior evaluation suites. To achieve this unified evaluation, UniEditBench standardizes prompt interfaces through a visually grounded triplet formulation (source prompt, target prompt, editing instruction), enabling direct evaluation across heterogeneous model input formats. As illustrated in Figure~\ref{fig:pipeline}, the full pipeline combines multi-source data aggregation, prompt unification, quality filtering, and evaluator distillation to form a consistent end-to-end benchmark workflow.

To support scalable and interpretable scoring within this workflow, we further build distilled MLLM-based evaluators by transferring visual assessment ability from a high-capacity teacher model (Qwen3-VL-235B-A22B) to smaller 4B/8B student models, initialized from Qwen3-VL-4B and Qwen3-VL-8B backbones. The resulting evaluators provide multi-dimensional assessment over structural fidelity, text alignment, background consistency, naturalness, and temporal-spatial consistency (for videos). Experiments show that these distilled evaluators maintain strong agreement with human judgments while substantially reducing deployment and inference costs relative to the teacher model. Overall, UniEditBench aims to provide a practical, reproducible, and accessible evaluation setup for the visual editing community.

\noindent\textbf{Contributions.} Our contributions are summarized as follows:
\begin{itemize}
\vspace{-2pt}
    \item \textbf{Unified benchmark across paradigms and modalities.} We introduce UniEditBench, which evaluates both image and video editing under a shared fine-grained taxonomy and protocol, enabling fair and consistent comparisons across previously isolated method families.
    
    \item \textbf{Benchmark construction with standardized prompt interfaces.} We build a multi-source, quality-controlled dataset and unify heterogeneous model inputs with visually grounded prompt triplets, improving evaluation consistency across different editing paradigms.
    
    \item \textbf{Cost-aware distilled evaluators with interpretable scoring.} We propose a teacher-student evaluation framework that distills a high-capacity MLLM judge into \textbf{lightweight} 4B/8B models and performs orthogonal multi-dimensional assessment, achieving strong alignment with human judgments at a fraction of the computational cost.
\end{itemize}

\begin{figure*}[!t]
    \centering
    \includegraphics[width=0.98\textwidth]{1_baseline1.jpg.pdf}
    \caption{Visual results of selected image editing methods on UniEditBench. See Appendix \ref{appendix:visualizations} for more examples.}
    \label{fig:qualitative_image}
\end{figure*}
\vspace{-0.5em}

\vspace{-3pt}
\section{Related Work}
\vspace{-1pt}

\subsection{Visual Generation and Editing}
Recent advances in visual content generation have been driven by diffusion models~\cite{ho_ddpm, rombach_stable_diffusion} and multimodal large language models (MLLMs)~\cite{liu_llava, qwen_vl_series}. Building on these foundations, research has shifted from holistic synthesis to controllable and fine-grained visual editing, enabling tasks such as object manipulation, attribute editing, and video-level transformations~\cite{hertz_prompt2prompt, brooks_instructpix2pix, zhang_controlnet}. 
Existing methods can be broadly grouped into two paradigms. Reconstruction-based approaches, including inversion-based~\cite{diffedit, rf_inversion} and inversion-free methods~\cite{flowedit, snr_edit}, emphasize structural preservation during targeted edits, with recent extensions to video settings. In contrast, instruction-driven methods use natural language or MLLM-based interfaces for flexible semantic modifications, as demonstrated by InstructPix2Pix~\cite{brooks_instructpix2pix}, Step1X-Edit~\cite{step1x}, Bagel~\cite{bagel}, and Qwen-Image~\cite{qwen_image}. 
These paradigms offer complementary strengths, but their differing assumptions and input formats complicate direct comparison and motivate a unified evaluation benchmark.

\vspace{-4pt}
\subsection{Evaluation Benchmarks for Visual Editing}
Evaluating visual editing quality remains challenging due to the complexity of disentangling intended edits from content preservation. Traditional metrics such as PSNR~\cite{PSNR} and CLIP Score~\cite{clipscore} often show limited alignment with human perception. To address this, several task-specific benchmarks, including EditVal~\cite{editval}, PIE-Bench, and EditBench~\cite{editbench}, have been proposed. 
However, these benchmarks typically focus on a single modality or a specific editing paradigm, and do not fully support unified evaluation across diverse methods. In addition, they provide limited coverage of compositional and reasoning-intensive tasks, such as object counting and spatial reordering. Extending evaluation to video editing further introduces challenges, as it requires jointly assessing temporal coherence and edit consistency across frames. Existing efforts, such as TGVE~\cite{tgve_benchmark}, provide initial exploration but remain limited in scope.
In contrast, UniEditBench introduces a unified evaluation protocol that spans both image and video editing, covering diverse editing operations under a consistent framework.

\vspace{-4pt}
\subsection{MLLM-based Evaluation and Distillation}
Recently, large language models serve as automated evaluators (LLM-as-a-Judge)~\cite{llm_as_a_judge}, demonstrating strong alignment with human judgments. In the visual domain, MLLMs enhance this by capturing fine-grained semantic and perceptual attributes beyond traditional metrics. 
However, directly employing large-scale MLLMs introduces practical limitations. Proprietary APIs incur substantial financial costs and privacy concerns, while deploying large open-source models like Qwen3-VL-235B-A22B~\cite{qwen_vl_series} requires significant computational resources. Knowledge distillation (KD)~\cite{mini_llm_kd, distilling} offers a principled approach to transferring capabilities from large teacher models to smaller student models.
Building on this idea, we develop a distilled MLLM-based evaluation framework that transfers capabilities from a high-capacity model to lightweight variants (4B/8B), enabling scalable, efficient evaluation for visual editing.

\vspace{-2pt}
\section{UniEditBench}
\vspace{-1pt}
We present the UniEditBench framework, designed for comprehensive visual editing assessment, by detailing the multi-source data collection and prompt unification pipeline in Section \ref{sec:methodology}. We outline the task taxonomy and dataset composition (Section \ref{sec:taxonomy}), followed by the formalization of our multi-dimensional evaluation metrics (Section \ref{sec:metrics}). Finally, Section \ref{sec:evaluator} introduces our cost-effective evaluator based on a two-stage knowledge distillation strategy.

\vspace{-2pt}
\subsection{Benchmark Construction Methodology}
\vspace{-1pt}
\label{sec:methodology}

To ensure diverse and high-fidelity source data, UniEditBench employs a multi-source aggregation pipeline spanning three domains (Figure~\ref{fig:pipeline}A). First, we sample from established benchmarks like PIE-Bench and ImgEdit \cite{imgedit} to enable backward comparability. Second, we crawl high-resolution internet images and videos to capture real-world complexities, such as natural motion, dynamic lighting, and cluttered backgrounds, which are often underrepresented in synthetic corpora. Third, we incorporate recent generative distributions via a taxonomy-driven synthesis pipeline. Specifically, we use GPT-5.4 to generate and filter 20 high-quality prompts per category, then deploy foundation models (FLUX \cite{flux} and SD3 \cite{sd3} for images; Wan \cite{wan} and HunyuanVideo \cite{hunyuanvideo} for videos) to synthesize diverse variations. This hybrid strategy systematically increases scene diversity while preserving controllable category coverage.

\vspace{0.3em}
\noindent\textbf{Visually-Grounded Dual-Paradigm Prompt Unification:} 
A key design goal is to \textbf{seamlessly} bridge reconstruction-based and instruction-driven evaluation with one consistent input interface (Figure~\ref{fig:pipeline}B). Instead of plain text rewriting, we use MLLMs to inspect underlying source media semantics and assign a fine-grained task label from our taxonomy. Conditioned on the task label and visual context, the system generates a structurally aligned prompt triplet: source prompt, target prompt, and editing instruction. This triplet allows direct adaptation across different editing paradigms while strictly preserving semantic equivalence. To reduce labeling noise, we apply multi-stage consistency checks, including secondary MLLM verification, strict rejection of ambiguous samples, and human-in-the-loop expert review for final acceptance.

\vspace{0.2em}
\noindent\textbf{Rigorous Quality Control:} 
The candidate pool undergoes a strict two-stage purification process (Figure~\ref{fig:pipeline}B). First, coarse-grained MLLM filtering removes samples with severe artifacts, watermarks, or poor text-media alignment. Second, expert annotators manually cross-validate the remaining assets to confirm edit feasibility, resolve prompt ambiguity, and rigorously verify temporal consistency for videos. We retain only samples that satisfy both semantic correctness and stringent visual quality criteria, thereby ensuring robust benchmark reliability for cross-model comparison.

\subsection{Taxonomy and Dataset Composition}
\label{sec:taxonomy}

To systematically evaluate visual editing models, UniEditBench constructs a highly granular taxonomy. Image editing tasks are categorized into 9 distinct operations: Add, Remove, Replace, Change, Stroke-based, Extract, Adjust, Count, and Reorder. To accommodate temporal dynamics, video editing is adapted into 8 core operations, excluding Stroke-based editing.

\vspace{0.2em}
\noindent\textbf{Beyond Appearance: The Challenge of Spatial Reasoning.} 
Traditional benchmarks primarily focus on appearance-level modifications (color alterations, style transfers) or simple entity swaps, often neglecting deep compositional understanding. UniEditBench addresses this by organizing its taxonomy along three functional axes: content modification (Add, Remove, Replace), attribute transformation (Change, Adjust, Extract), and compositional reasoning (Count, Reorder). The explicitly introduced Count task demands precise numerical-spatial mapping, adjusting object quantities without altering the surrounding unedited backgrounds. Similarly, the Reorder task requires complex spatial logic and object permanence, requiring subjects to be moved within a frame while maintaining overall scene visual consistency. These advanced subsets directly target existing generative models' weaknesses in complex spatial reasoning and multi-object relation alignment.

\begin{figure}[!b]
    \centering
    \includegraphics[width=\linewidth]{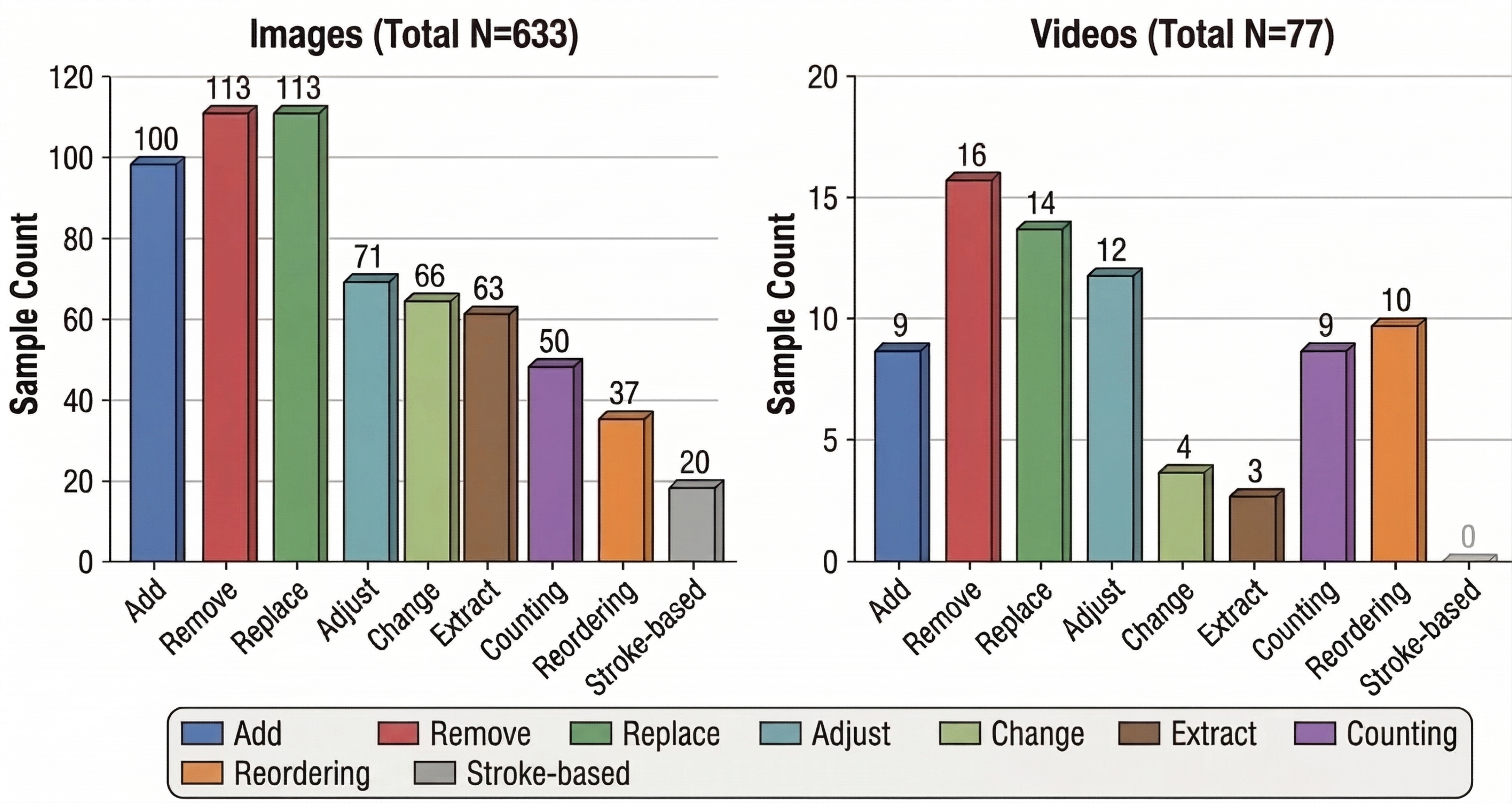}
    \vspace{-2pt}
    \caption{Data distribution of UniEditBench samples across various editing tasks for image and video modalities.}
    \label{fig:111}
    \vspace{-2pt}
\end{figure}

\vspace{0.5em}
\noindent\textbf{Dataset Statistics and Domain Diversity.} 
The UniEditBench dataset comprises 633 high-quality source images and 77 source videos, yielding 710 precisely aligned prompt triplets. Figure \ref{fig:111} illustrates the detailed distribution of these samples across the defined image and video editing tasks. To guarantee robustness, the visual assets are curated for broad domain diversity, encompassing a wide spectrum of artistic styles (realistic photography, 2D anime, 3D rendering, oil painting) and rich semantic scenes (portraits, animals, landscapes, urban environments).

\subsection{Multi-Dimensional Orthogonal Metrics}
\label{sec:metrics}

Traditional holistic or scalar metrics often suffer from entanglement, where a localized error (e.g., background distortion) inadvertently lowers the score for unrelated aspects (e.g., instruction following). To mitigate this, UniEditBench introduces a decoupled, multi-dimensional orthogonal evaluation system featuring five independent dimensions, each scored on a 1 to 5 Likert scale:

\begin{itemize}
    \item Structural Fidelity ($S_{SF}$): Evaluates the overall structural and spatial preservation of the edited subject or scene. It explicitly penalizes unintended morphological changes, ensuring the fundamental geometry and layout remain completely intact despite semantic alterations.
    
    \item Text Alignment ($S_{TA}$): Quantifies the strict semantic adherence to the given target prompt or editing instruction, comprehensively assessing whether the requested operation is accurately executed on the intended subject.
    
    \item Background Consistency ($S_{BC}$): Explicitly isolates unedited regions to measure the model's ability to preserve background pixels and context perfectly, penalizing over-editing or unintended color bleeding.
    
    \item Naturalness ($S_{NAT}$): Evaluates the overall perceptual realism and aesthetic visual quality completely independent of the prompt, focusing on generative artifacts, unnatural blending boundaries, and irrational lighting conditions.
    
    \item Temporal-Spatial Consistency ($S_{TSC}$, Video Only): Tailored specifically for the video domain to rigorously assess overall cross-frame coherence, strictly penalizing visual flickering, jittering, and discontinuous motion trajectories.
\end{itemize}

This decoupled evaluation reduces semantic overlap, facilitating interpretable failure analysis and revealing inherent trade-offs across different model architectures.

\subsection{Cost-Effective Evaluator via KD}
\label{sec:evaluator}

Deploying flagship MLLMs for automated evaluation remains inaccessible to the broader research community despite achieving high human alignment. Proprietary models (GPT-5.4, Gemini 3.1 Pro) incur prohibitive financial costs when processing large-scale benchmarks via commercial APIs. Conversely, flagship open-source models (Qwen3-VL-235B-A22B) require substantial local infrastructure—typically an 8-way A100 GPU cluster—exceeding the capacity of standard academic laboratories. To democratize standardized evaluation, we distill the visual reasoning and discriminative capabilities of Qwen3-VL-235B-A22B into efficient 4B and 8B student evaluators based on Qwen3-VL-4B and Qwen3-VL-8B.

\vspace{0.5em}
\noindent\textbf{Teacher Data Generation via Chain-of-Thought(CoT):} 
To construct a robust preference dataset, we generate edited instances across the entire benchmark using a diverse suite of representative models spanning different editing paradigms. Image editing frameworks include Bagel \cite{bagel}, Step1X-Edit \cite{step1x}, Qwen-Image-Edit \cite{qwen_image}, RF-Inversion \cite{rf_inversion}, and FlowEdit \cite{flowedit}. Video models include FlowEdit-Wan \cite{floweditvideo}, FlowDirector \cite{flowdirector}, InstructV2V \cite{instructv2v}, and Ditto \cite{ditto}. We then deploy the Qwen3-VL-235B-A22B teacher model to score these results. To ensure the teacher generates rationalized data rather than mere scalar values, we design a CoT prompting strategy, with the precise instructional templates detailed in Appendix \ref{appendix:prompts}. This mechanism compels the model to generate a detailed semantic rationale for each of the five orthogonal dimensions (Section \ref{sec:metrics}) before outputting the final discrete Likert scores, creating a rich multimodal reasoning dataset for student emulation.

\vspace{0.5em}
\noindent\textbf{Curriculum-based Two-Stage LoRA Distillation:} 
To transfer the teacher's discriminative logic efficiently, we utilize the ms-swift framework \cite{msswift} for LoRA fine-tuning. Student models are trained to minimize the distribution discrepancy between their predictions and the teacher's multi-dimensional scores. Formally, for each evaluation dimension $k$, the optimization objective is defined as:
$$ \mathcal{L}_{distill} = \sum_{k} \mathbb{E} [ \| S_k^{teacher} - S_k^{student} \|^2 ] $$
where $S_k$ denotes the score for a specific dimension. To prevent catastrophic forgetting of spatial reasoning when introducing temporal dynamics, we implement a two-stage curriculum learning strategy. In Stage 1 (Spatial Distillation), students are fine-tuned exclusively on the image evaluation dataset for 3 epochs, anchoring foundational capabilities in structural fidelity, text alignment, and background consistency. In Stage 2 (Temporal Distillation), models initialized with Stage 1 weights are fine-tuned on the video dataset for another 3 epochs. This progressive approach injects the ability to assess Temporal-Spatial Consistency while preserving spatial metric accuracy.

\vspace{0.5em}
\noindent\textbf{Validation and Computational Efficiency:} 
To quantitatively validate our distillation paradigm, we monitor the Mean Squared Error (MSE) against the teacher's gold-standard scores (Table \ref{tab:mse_operations}). Post-fine-tuning, both models exhibit significant MSE reductions, yet they demonstrate distinct and complementary strengths. The 8B variant holds a clear advantage in high-difficulty, reasoning-intensive tasks including Extract, Reorder, Replace, and Count. Conversely, the 4B model maintains robust comprehensive capabilities, striking an optimal performance balance across both image and video evaluation modalities. Beyond technical alignment, we further validate these evaluators by confirming their high correlation with human preference judgments. Crucially, as detailed in Table \ref{tab:qwen3vl-specs}, this framework achieves a drastic reduction in computational overhead, shifting the hardware requirement from a massive multi-GPU cluster to a single consumer-grade GPU and bypassing expensive API dependencies.
\vspace{-0.5em}

\section{Experiment}
\subsection{Experimental Settings}
\label{sec:exp_setting}

\vspace{0.3em}
\noindent\textbf{Benchmark Baselines.} We evaluate 25 open-source image editing models (spanning inversion-based, inversion-free, structured, MLLM-based, masked diffusion, and autoregressive approaches) and 8 open-source video editing models. This diverse selection ensures broad paradigm coverage and reproducible comparisons. By including methods with varying strengths in reconstruction fidelity, instruction following, and semantic flexibility, we can effectively evaluate whether our benchmark fairly exposes inherent model trade-offs without structural bias.

\begin{figure*}[!t]
    \centering
    \includegraphics[width=\textwidth]{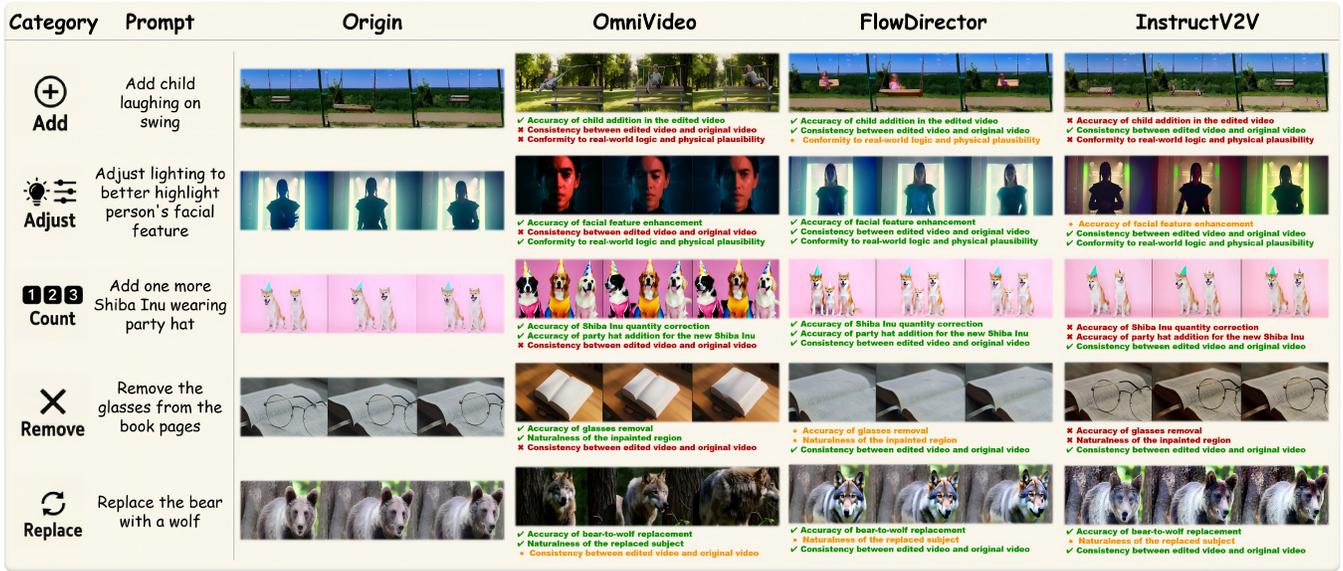}
    \caption{Visual results of selected video editing methods on UniEditBench. See Appendix \ref{appendix:visualizations} for more examples.}
    \label{fig:qualitative_vid}
\end{figure*}

\vspace{0.3em}
\noindent\textbf{Evaluator Configurations.} We use the two-stage distilled Qwen3-VL-4B and Qwen3-VL-8B as our primary evaluators. They score image models across four orthogonal dimensions ($S_{SF}$, $S_{TA}$, $S_{BC}$, $S_{NAT}$) and video models across five (adding $S_{TSC}$). All results are reported in a 4B/8B dual-entry format. To measure distillation accuracy, we calculate the Mean Squared Error (MSE) against the Qwen3-VL-235B-A22B teacher scores. Category-specific MSE is averaged across samples within each editing operation, while the Overall MSE represents the global error across the entire image or video dataset. Applying this unified scoring space across all paradigms ensures that cross-method comparisons remain consistent and interpretable.

\vspace{0.3em}
\noindent\textbf{Human Alignment Study.} To validate evaluator reliability, we conducted a blind user study with 50 participants, comprising both academic experts and general users. Participants were divided into five groups to independently evaluate the entire UniEditBench dataset without sample overlap, ensuring five complete evaluation passes. We selected representative models across distinct paradigms (RF\_Inversion, FlowEdit, Step1X-Edit, and Bagel\_with\_thinking for images; FlowEdit-Wan, ICVE, and OmniVideo for videos) to establish ground-truth human preference scores. These reference scores allow us to verify if our distilled evaluators consistently align with human judgments across varying editing tasks.
\vspace{-0.5em}

% --- Table 1: Image Main Benchmark ---
\begin{table*}[!htbp]
\centering
\caption{Quantitative Benchmarking on \textbf{Image Editing}. Scores (1-5) are evaluated by our distilled SFT Image+Video evaluators. Each cell presents assessment results in a \textbf{4B/8B} dual-entry format. Best results within each paradigm are \textbf{bolded}, and second-best results are \underline{underlined}. The table highlights why unified, multi-dimensional evaluation is necessary: different paradigms excel on different dimensions, and no single method dominates all criteria simultaneously.}
\vspace{-0.2em}
\label{tab:main_image}
\renewcommand{\arraystretch}{1}
\resizebox{\textwidth}{!}{
\begin{tabular}{llccccc}
\toprule
\textbf{Paradigm} & \textbf{Model} & \textbf{Structural ($S_{SF}$)} & \textbf{Alignment ($S_{TA}$)} & \textbf{Background ($S_{BC}$)} & \textbf{Naturalness ($S_{NAT}$)} & \textbf{Overall} \\
\midrule
\multirow{5}{*}{Inversion-based} 
 & SDEdit\cite{sdedit} & 1.74/2.12 & \underline{2.60/2.78} & 1.49/1.70 & 1.72/1.99 & 1.89/2.15 \\
 & RF\_Inversion\cite{rf_inversion} & 2.69/2.64 & 2.11/2.09 & 2.36/2.32 & 2.33/2.27 & 2.37/2.33 \\
 & RF\_Solver\cite{rf_solver} & 3.26/3.29 & \textbf{3.33/3.32} & 2.52/2.61 & 2.61/2.69 & 2.93/2.98 \\
 & PNP\_Inversion\cite{pnpinversion} & \underline{3.63/3.68} & 2.51/2.50 & \underline{3.10/3.17} & \underline{2.70/2.82} & \underline{2.99/3.04} \\
 & Masactrl\cite{masactrl} & \textbf{3.82/3.86} & 2.37/2.33 & \textbf{3.67/3.72} & \textbf{2.95/3.04} & \textbf{3.20/3.24} \\
\midrule
\multirow{3}{*}{Inversion-free}
 & Flowedit\cite{flowedit} & \underline{3.84/3.79} & \textbf{3.36/3.33} & \underline{3.96/3.92} & \underline{3.38/3.36} & \underline{3.63/3.60} \\
 & FTEdit\cite{fiaedit} & 3.61/3.65 & \underline{3.18/3.13} & 3.37/3.50 & 2.98/3.06 & 3.28/3.33 \\
 & DNAEdit\cite{dnaedit} & \textbf{4.40/4.40} & 2.64/2.58 & \textbf{4.37/4.41} & \textbf{3.76/3.79} & \textbf{3.79/3.80} \\
\midrule
\multirow{4}{*}{Structured editing}
 & InstructPix2Pix\cite{instructpix2pix} & 1.65/1.74 & 1.65/1.79 & 1.34/1.39 & 1.23/1.29 & 1.47/1.55 \\
 & Step1X-Edit\cite{step1x} & \textbf{4.53/4.47} & \textbf{3.88/3.87} & \textbf{4.58/4.58} & \textbf{4.01/3.94} & \textbf{4.25/4.21} \\
 & ICEdit\cite{ICEdit} & \underline{4.29/4.25} & \underline{3.08/3.01} & \underline{4.26/4.21} & \underline{3.60/3.58} & \underline{3.81/3.76} \\
 & control\_v11e\_sd15\_ip2p\cite{controlnet} & 1.64/1.62 & 1.66/1.80 & 1.18/1.21 & 1.07/1.09 & 1.39/1.43 \\
\midrule
\multirow{7}{*}{MLLM}
 & Hunyuan\cite{hunyuanimage} & \textbf{4.62/4.69} & \textbf{4.63/4.61} & 4.10/4.16 & \textbf{4.13/4.15} & \textbf{4.37/4.40} \\
 & Qwen-Image-edit\cite{qwen_image} & 4.54/4.53 & \underline{4.36/4.35} & 4.17/4.18 & \underline{4.09/4.09} & \underline{4.29/4.29} \\
 & Bagel\_with\_thinking\cite{bagel} & 4.20/4.14 & 3.64/3.64 & 4.29/4.34 & 3.52/3.52 & 3.91/3.91 \\
 & Bagel\_without\_thinking\cite{bagel} & 4.17/4.17 & 3.71/3.70 & \textbf{4.37/4.38} & 3.60/3.58 & 3.96/3.96 \\
 & UniWorld\cite{uniworld} & 4.24/4.30 & 3.85/3.85 & 3.98/4.13 & 3.69/3.78 & 3.94/4.02 \\
 & Omni-Gen2\cite{omnigen2} & 4.39/4.38 & 3.65/3.60 & 4.07/4.17 & 3.71/3.79 & 3.96/3.99 \\
 & DreamOmni2\cite{dreamomni2} & \underline{4.53/4.56} & 3.54/3.48 & \underline{4.33/4.33} & 4.04/4.02 & 4.11/4.10 \\
\midrule
\multirow{4}{*}{Masked Diff/AR}
 & EditMGT\cite{editmgt} & 2.59/2.61 & 2.67/2.75 & 2.83/2.91 & 1.89/1.94 & 2.49/2.55 \\
 & LaVida-O\cite{lavida} & \textbf{3.85/3.89} & \textbf{3.54/3.59} & \underline{3.49/3.55} & \underline{2.88/2.96} & \underline{3.44/3.50} \\
 & EditAR\cite{editar} & 3.29/3.37 & 2.56/2.55 & 3.03/3.07 & 2.50/2.60 & 2.85/2.90 \\
 & VAREdit\cite{VAREdit} & \underline{3.84/3.82} & \underline{3.53/3.55} & \textbf{4.21/4.29} & \textbf{3.50/3.51} & \textbf{3.77/3.79} \\
\midrule
\multirow{2}{*}{Others}
 & AnyEdit\cite{anyedit} & \underline{3.61/3.62} & \underline{2.46/2.43} & \underline{3.80/3.82} & \underline{2.81/2.82} & \underline{3.17/3.17} \\
 & FLUX.1-Kontext-dev\cite{flux-kontext} & \textbf{4.62/4.63} & \textbf{3.74/3.71} & \textbf{4.63/4.64} & \textbf{4.25/4.22} & \textbf{4.31/4.30} \\
\bottomrule
\end{tabular}
}
\end{table*}

% --- Table 2: Video Main Benchmark ---
\begin{table*}[!htbp]
\centering
\caption{Quantitative Benchmarking on \textbf{Video Editing}. Scores (1-5) are evaluated by our distilled SFT Image+Video evaluators. Each cell presents assessment results in a \textbf{4B/8B} dual-entry format. Best results within each paradigm are \textbf{bolded}, and second-best results are \underline{underlined}. The results emphasize our motivation that video editing must be evaluated jointly on instruction following and temporal stability, rather than by single-aspect quality alone.}
\vspace{-0.2em}
\label{tab:main_video}
\renewcommand{\arraystretch}{1}
\resizebox{\textwidth}{!}{
\begin{tabular}{llcccccc}
\toprule
\textbf{Paradigm} & \textbf{Model} & \textbf{Structural ($S_{SF}$)} & \textbf{Alignment ($S_{TA}$)} & \textbf{Background ($S_{BC}$)} & \textbf{Naturalness ($S_{NAT}$)} & \textbf{Temporal ($S_{TSC}$)} & \textbf{Overall} \\
\midrule
\multirow{3}{*}{Inversion-free}
 & FlowDirector\cite{flowdirector} & \underline{3.96/4.10} & \textbf{3.87/3.87} & \underline{4.68/4.57} & \underline{4.13/4.04} & \underline{4.39/4.35} & \underline{4.21/4.19} \\
 & FlowEdit-Pyramid\cite{floweditvideo} & 3.77/3.56 & \underline{3.55/3.42} & 4.04/3.86 & 3.45/3.19 & 3.69/3.39 & 3.70/3.48 \\
 & FlowEdit-Wan\cite{floweditvideo} & \textbf{4.25/4.40} & 3.30/3.31 & \textbf{4.74/4.74} & \textbf{4.47/4.48} & \textbf{4.66/4.69} & \textbf{4.28/4.32} \\
\midrule
\multirow{2}{*}{Structured editing}
 & ICVE\cite{icve} & \underline{3.97/3.90} & \textbf{4.00/3.94} & \textbf{4.45/4.43} & \textbf{3.70/3.70} & \textbf{4.08/4.00} & \textbf{4.04/3.99} \\
 & InstructV2V\cite{instructv2v} & \textbf{4.01/4.00} & \underline{3.38/3.31} & \underline{3.90/3.84} & \underline{3.55/3.61} & \underline{3.86/3.90} & \underline{3.74/3.73} \\
\midrule
\multirow{2}{*}{MLLM}
 & OmniVideo\cite{omnivideo} & \underline{0.40/0.40} & \underline{1.29/1.29} & \underline{0.45/0.60} & \underline{0.83/0.86} & \underline{0.57/0.61} & \underline{0.71/0.75} \\
 & Ditto\cite{ditto} & \textbf{2.99/2.94} & \textbf{3.19/3.21} & \textbf{2.97/2.75} & \textbf{2.78/2.56} & \textbf{2.97/2.84} & \textbf{2.98/2.86} \\
\midrule
Others & Lucy-Edit & 3.42/3.44 & 3.17/3.14 & 4.05/4.09 & 3.30/3.27 & 3.51/3.52 & 3.49/3.49 \\
\bottomrule
\end{tabular}
}
\end{table*}

\vspace{-0.5em}
\subsection{Main Results and Analysis}
\label{sec:main_results}

\noindent\textbf{Quantitative Analysis on Image Editing.} Table \ref{tab:main_image} reports model performance on four orthogonal dimensions ($S_{SF}$, $S_{TA}$, $S_{BC}$, $S_{NAT}$). Overall, MLLM-based methods generally obtain higher text-alignment and naturalness scores, consistent with their stronger semantic priors. Within this group, Hunyuan and Qwen-Image-edit achieve high overall averages, while different models exhibit distinct trade-offs between instruction following and background preservation. Structured editing methods present another strong frontier: Step1X-Edit performs particularly well on structural fidelity and keeps competitive overall scores. Inversion-free approaches (e.g., FlowEdit, DNAEdit) remain competitive on structural and background-related dimensions, suggesting advantages in preserving scene layout during editing. Inversion-based methods, by contrast, often preserve global context but show weaker performance on complex instruction execution. These trends indicate that no single paradigm dominates all dimensions, reinforcing the value of orthogonal evaluation. Representative qualitative examples are shown in Figure \ref{fig:qualitative_image}.

\vspace{0.3em}
\noindent\textbf{Quantitative Analysis on Video Editing.} Video editing is more challenging because models must satisfy both edit correctness and temporal stability (Table \ref{tab:main_video}). FlowEdit-Wan achieves strong performance on temporal-spatial consistency and naturalness, while ICVE shows strong alignment scores but can exhibit temporal instability in some cases. This trade-off pattern highlights that text alignment alone is insufficient for practical video editing quality; temporal coherence must be jointly assessed. The results also show larger performance gaps between methods than in the image setting, suggesting video editing remains a less mature and more variable regime. Qualitative comparisons in Figure \ref{fig:qualitative_vid} further corroborate these quantitative observations.

\begin{figure}[!htbp]
    \centering
    \includegraphics[width=\linewidth]{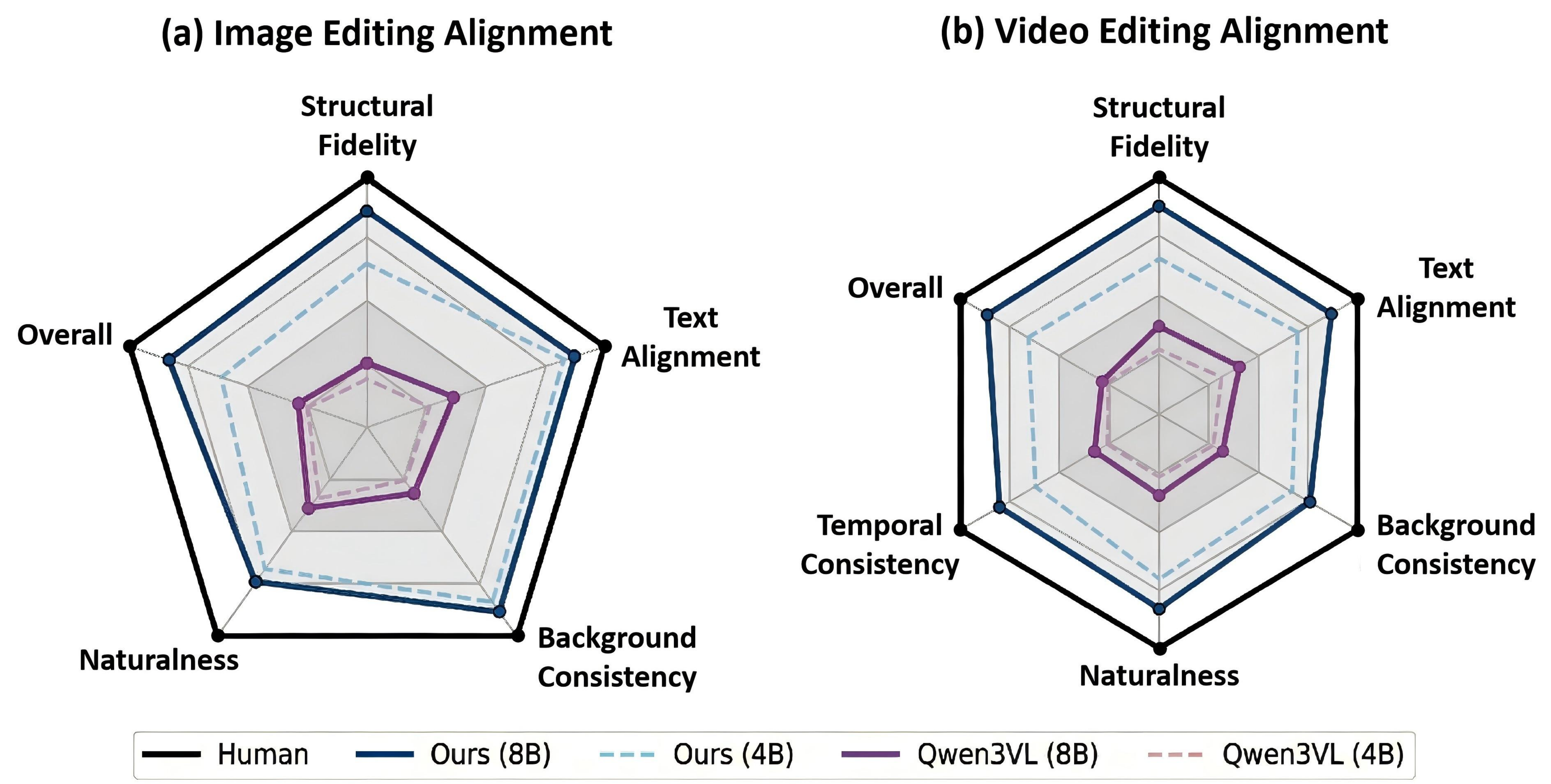}
    \caption{Radar chart comparing alignment of automated evaluators with human preference across multiple dimensions. Distilled models demonstrate significantly higher correlation than zero-shot baselines of equivalent scale.}
    \label{fig:radar}
    \vspace{-8pt}
\end{figure}

\vspace{0.3em}
\noindent\textbf{Human Preference Alignment.} As shown in Figure \ref{fig:radar}, the distilled evaluators exhibit strong alignment with human preference. Compared with zero-shot models of similar scale, the fine-tuned evaluators better match human scoring trends in this benchmark setting, supporting their use as practical automated judges. This result is central to our motivation: scalable evaluation is only useful if it remains perceptually meaningful. By narrowing the gap between automatic scores and human judgment, the distilled evaluators make large-scale, repeatable benchmarking more feasible for both image and video editing research.
\vspace{-0.5em}

% --- Table 3: MSE by Operations ---
\begin{table*}[!htbp]
\centering
\caption{Mean Squared Error (MSE) compared to the Qwen3-VL-235B-A22B teacher model, categorized by editing operations. Values are calculated as a weighted aggregate based on sample distributions. Our two-stage SFT models achieve the lowest MSE across all tasks. More detailed supporting experimental data are provided in Appendix \ref{appendix:detailed_mse}.}
\label{tab:mse_operations}
\renewcommand{\arraystretch}{1.1}
\resizebox{\textwidth}{!}{
\begin{tabular}{l|ccccccccc|c|c}
\toprule
\textbf{Model} & \textbf{Add} & \textbf{Adjust} & \textbf{Change} & \textbf{Count} & \textbf{Extract} & \textbf{Remove} & \textbf{Reorder} & \textbf{Replace} & \textbf{Stroke-based} &\textbf{Ovr. (Image)} & \textbf{Ovr. (Video)} \\
\midrule
Zero-shot 4B  & 2.16 & 2.27 & 3.71 & 3.87 & 5.98 & 3.62 & 3.73 & 2.47 & 1.47 & 2.95 & 5.40 \\
Zero-shot 8B  & 1.50 & 1.58 & 2.73 & 2.83 & 5.24 & 3.25 & 3.84 & 2.03 & 2.11 & 2.51 & 4.00 \\
\midrule
SFT Image 4B  & 0.35 & 0.46 & 0.45 & 0.76 & 1.62 & 1.12 & 0.95 & 0.43 & 0.32 & 0.50 & 2.33 \\
SFT Image 8B  & 0.27 & 0.31 & 0.56 & 0.60 & 1.15 & 0.93 & 0.87 & 0.40 & 0.31 & 0.47 & 1.56 \\
\midrule
\textbf{SFT Image+Video 4B} & \textbf{0.27} & \textbf{0.25} & \textbf{0.39} & \textbf{0.48} & \textbf{1.30} & \textbf{0.72} & \textbf{0.67} & \textbf{0.36} & \textbf{0.31} & \textbf{0.49} & \textbf{0.73} \\
\textbf{SFT Image+Video 8B} & \textbf{0.23} & \textbf{0.26} & \textbf{0.40} & \textbf{0.46} & \textbf{1.24} & \textbf{0.78} & \textbf{0.61} & \textbf{0.31} & \textbf{0.32} & \textbf{0.47} & \textbf{0.79} \\
\bottomrule
\end{tabular}
}
\vspace{0.2cm}
\end{table*}

\vspace{0.2em}
% --- Table 4: Deployment Efficiency ---
\begin{table*}[!htbp]
\centering
\caption{Deployment \& Inference Specifications for Evaluators (BF16 Precision)}
\vspace{-0.2em}
\label{tab:qwen3vl-specs}
\begin{tabular}{lccc}
\toprule
\textbf{Feature / Model Scale} & \textbf{Our SFT 4B} & \textbf{Our SFT 8B} & \textbf{Qwen3-VL-235B-A22B} \\ 
\midrule
Architecture Type & Dense & Dense & Mixture of Experts (MoE) \\
Total Parameters & 4B & 8B & 235B \\
Activated Params (per Token) & 4B & 8B & 22B \\
Weight VRAM (BF16) & $\approx 8.8$ GB & $\approx 17.2$ GB & $\approx 470.8$ GB \\
1K Image Inference (Extra VRAM) & $\approx 1.2$ GB & $\approx 2.4$ GB & $\approx 12-18$ GB \\
Suggested Minimum VRAM & 12 GB & 24 GB & 640 GB (8$\times$80GB Cluster) \\
Avg. Inference Time (1K Image) & 0.6 - 1.0 s & 1.0 - 1.5 s & 5.0 - 7.0 s (8 GPUs) \\
\bottomrule
\end{tabular}
\end{table*}

\vspace{-0.2em}
\subsection{Effectiveness of Knowledge Distillation}
\label{sec:distillation}

To validate our knowledge distillation strategy, we measure the MSE between our evaluators and the 235B teacher (Table \ref{tab:mse_operations}). More detailed experimental results are provided in Appendix \ref{appendix:detailed_mse}.

\vspace{0.3em}
\noindent\textbf{Necessity of Two-Stage SFT.} Single-stage (image-only) fine-tuning sharply reduces MSE on image tasks but suffers a severe error rebound when evaluated on video metrics. Introducing the second temporal stage resolves this disparity, confirming that our curriculum learning successfully injects temporal knowledge without catastrophic forgetting of spatial reasoning.

\vspace{0.3em}
\noindent\textbf{Scale vs. Precision.} While both SFT models vastly outperform zero-shot baselines, the 8B variant consistently achieves lower MSE on reasoning-heavy tasks (Count, Reorder, Extract). This indicates larger parameter capacities better internalize the teacher's discriminative logic, though the 4B model remains a highly reliable, lightweight alternative.

\vspace{-0.2em}
\subsection{Computational and Deployment Costs}
\label{sec:efficiency}

We compare deployment specifications of the distilled evaluators and the 235B teacher model in Table \ref{tab:qwen3vl-specs}. The teacher model requires substantially larger BF16 weight storage ($\approx 470.8$ GB) and multi-GPU infrastructure, while the distilled dense evaluators can run with much lower VRAM requirements (12 GB for 4B, 24 GB for 8B). In our measurements, this reduction in model scale is accompanied by notably lower single-image inference latency. These encouraging results indicate that our distillation process can significantly improve evaluation throughput and accessibility while retaining strong agreement with teacher and human judgments.

\vspace{1.2em}
\section{Conclusion}

\vspace{0.6em}

We present \textbf{UniEditBench}, a unified benchmark for evaluating both image and video editing models. UniEditBench provides a shared protocol across reconstruction-based and instruction-driven paradigms, a taxonomy that includes compositional tasks such as counting and reordering, and a multi-dimensional evaluation setup for interpretable analysis. This combination addresses three practical bottlenecks at once: fragmented cross-paradigm evaluation, incomplete video editing evaluation systems, and the gap between low-cost metrics and human perception. To improve evaluation practicality, we introduce distilled 4B/8B evaluators trained from a high-capacity MLLM teacher. Experimental results show strong agreement with human judgments and meaningful reductions in deployment requirements relative to the teacher model. Together, these design choices make UniEditBench both methodologically rigorous and practically deployable, and we hope it can serve as a reproducible, accessible basis for future editing research.

\vspace{0.6em}
\noindent\textbf{Limitations and Future Work:} The current benchmark primarily targets single-turn editing. Many real-world workflows are iterative and conversational, requiring models to preserve context over multiple rounds of edits. Extending UniEditBench to multi-turn image and video editing trajectories is an important next step. In addition, while distilled evaluators improve scalability, they may still inherit biases from the teacher model; future work can explore stronger bias diagnosis and cross-evaluator calibration. Another limitation is that our taxonomy and sample scale, while broad, cannot fully cover long-tail real-world editing intents (e.g., culturally specific semantics, rare object interactions, and highly ambiguous instructions), which may affect external validity. 
%Finally, because the evaluation dimensions and prompt interfaces are fixed, strong methods may partially overfit to benchmark-specific priors; future releases should incorporate continual data refresh and adversarial stress tests to improve robustness.


\bibliographystyle{ACM-Reference-Format}
\bibliography{sample-sigconf-xelatex}


\clearpage
\appendix

\section{Detailed Alignment Error Analysis}
\label{appendix:detailed_mse}

This appendix provides the detailed, unaggregated Mean Squared Error (MSE) data supporting the consolidated ablation results presented in the main text. While the main manuscript reports a weighted aggregate to concisely summarize overall alignment, the tables below present the exact MSE breakdowns strictly separated by image and video modalities. 

Tables \ref{tab:app_img_ops} and \ref{tab:app_vid_ops} detail the alignment errors categorized by fine-grained editing operations for the image and video datasets, respectively. To provide a deeper structural analysis, Tables \ref{tab:app_img_dims} and \ref{tab:app_vid_dims} further decouple these errors across the orthogonal evaluation dimensions defined in the benchmark. These comprehensive metrics confirm that the two-stage fine-tuned models consistently achieve the lowest divergence from the 235B teacher model across all isolated tasks and dimensions.

% --- Appendix Table 1: Image MSE by Operations ---
\begin{table*}[htbp]
\centering
\caption{Detailed MSE on \textbf{Image Editing} categorized by editing operations.}
\label{tab:app_img_ops}
\renewcommand{\arraystretch}{1.2}
\resizebox{\textwidth}{!}{
\begin{tabular}{lcccccccccc}
\toprule
\textbf{Model} & \textbf{Add} & \textbf{Adjust} & \textbf{Change} & \textbf{Count} & \textbf{Extract} & \textbf{Remove} & \textbf{Reorder} & \textbf{Replace} & \textbf{Stroke} & \textbf{Overall} \\
\midrule
Zero-shot 4B  & 1.94 & 1.91 & 3.43 & 2.85 & 5.92 & 3.38 & 4.00 & 2.11 & 1.47 & 2.95 \\
Zero-shot 8B  & 1.34 & 1.46 & 2.46 & 2.34 & 5.24 & 3.00 & 3.95 & 1.87 & 2.11 & 2.51 \\
\midrule
SFT Image 4B  & 0.28 & 0.28 & 0.33 & 0.48 & 1.15 & 0.72 & 0.63 & 0.34 & 0.32 & 0.50 \\
SFT Image 8B  & 0.24 & 0.23 & 0.34 & 0.47 & 1.01 & 0.71 & 0.57 & 0.33 & 0.31 & 0.47 \\
\midrule
SFT Image+Video 4B & 0.25 & 0.24 & 0.36 & 0.46 & 1.26 & 0.65 & 0.61 & 0.35 & 0.31 & 0.49 \\
SFT Image+Video 8B & 0.22 & 0.24 & 0.32 & 0.44 & 1.23 & 0.69 & 0.59 & 0.28 & 0.32 & 0.47 \\
\bottomrule
\end{tabular}
}
\vspace{0.6cm}
\end{table*}

% --- Appendix Table 2: Image MSE by Dimensions ---
\begin{table*}[htbp]
\centering
\caption{Detailed MSE on \textbf{Image Editing} categorized by orthogonal evaluation dimensions.}
\label{tab:app_img_dims}
\renewcommand{\arraystretch}{1.2}
\resizebox{0.8\textwidth}{!}{
\begin{tabular}{lccccc}
\toprule
\textbf{Model} & \textbf{Structural ($S_{SF}$)} & \textbf{Alignment ($S_{TA}$)} & \textbf{Background ($S_{BC}$)} & \textbf{Naturalness ($S_{NAT}$)} & \textbf{Overall} \\
\midrule
Zero-shot 4B  & 3.62 & 2.26 & 2.68 & 3.25 & 2.95 \\
Zero-shot 8B  & 2.73 & 2.40 & 2.35 & 2.55 & 2.51 \\
\midrule
SFT Image 4B  & 0.66 & 0.28 & 0.57 & 0.49 & 0.50 \\
SFT Image 8B  & 0.63 & 0.28 & 0.51 & 0.45 & 0.47 \\
\midrule
SFT Image+Video 4B & 0.62 & 0.29 & 0.57 & 0.48 & 0.49 \\
SFT Image+Video 8B & 0.57 & 0.28 & 0.57 & 0.47 & 0.47 \\
\bottomrule
\end{tabular}
}
\vspace{0.6cm}
\end{table*}

% --- Appendix Table 3: Video MSE by Operations ---
\begin{table*}[htbp]
\centering
\caption{Detailed MSE on \textbf{Video Editing} categorized by editing operations.}
\label{tab:app_vid_ops}
\renewcommand{\arraystretch}{1.2}
\resizebox{0.95\textwidth}{!}{
\begin{tabular}{lccccccccc}
\toprule
\textbf{Model} & \textbf{Add} & \textbf{Adjust} & \textbf{Change} & \textbf{Count} & \textbf{Extract} & \textbf{Remove} & \textbf{Reorder} & \textbf{Replace} & \textbf{Overall} \\
\midrule
Zero-shot 4B  & 4.43 & 4.02 & 8.14 & 8.50 & 7.28 & 5.04 & 3.01 & 6.16 & 5.40 \\
Zero-shot 8B  & 3.16 & 2.19 & 6.88 & 5.07 & 5.29 & 4.80 & 3.54 & 3.70 & 4.00 \\
\midrule
SFT Image 4B  & 1.03 & 1.35 & 2.36 & 2.01 & 10.95 & 3.53 & 1.83 & 1.37 & 2.33 \\
SFT Image 8B  & 0.60 & 0.68 & 3.93 & 1.19 & 3.85 & 2.25 & 1.69 & 1.08 & 1.56 \\
\midrule
SFT Image+Video 4B & 0.52 & 0.28 & 0.80 & 0.57 & 2.06 & 1.13 & 0.84 & 0.51 & 0.73 \\
SFT Image+Video 8B & 0.36 & 0.34 & 1.73 & 0.56 & 1.50 & 1.34 & 0.65 & 0.64 & 0.79 \\
\bottomrule
\end{tabular}
}
\vspace{0.6cm}
\end{table*}

% --- Appendix Table 4: Video MSE by Dimensions ---
\begin{table*}[htbp]
\centering
\caption{Detailed MSE on \textbf{Video Editing} categorized by orthogonal evaluation dimensions.}
\label{tab:app_vid_dims}
\renewcommand{\arraystretch}{1.2}
\resizebox{\textwidth}{!}{
\begin{tabular}{lcccccc}
\toprule
\textbf{Model} & \textbf{Structural ($S_{SF}$)} & \textbf{Alignment ($S_{TA}$)} & \textbf{Background ($S_{BC}$)} & \textbf{Naturalness ($S_{NAT}$)} & \textbf{Temporal ($S_{TSC}$)} & \textbf{Overall} \\
\midrule
Zero-shot 4B  & 7.03 & 4.61 & 4.58 & 4.77 & 6.01 & 5.40 \\
Zero-shot 8B  & 5.66 & 3.81 & 2.49 & 3.40 & 4.63 & 4.00 \\
\midrule
SFT Image 4B  & 3.42 & 1.67 & 2.08 & 2.07 & 2.43 & 2.33 \\
SFT Image 8B  & 2.06 & 1.53 & 1.14 & 1.39 & 1.65 & 1.56 \\
\midrule
SFT Image+Video 4B & 0.87 & 0.80 & 0.58 & 0.65 & 0.75 & 0.73 \\
SFT Image+Video 8B & 1.00 & 0.83 & 0.60 & 0.69 & 0.81 & 0.79 \\
\bottomrule
\end{tabular}
}
\end{table*}

\section{Prompt Templates for Knowledge Distillation}
\label{appendix:prompts}

To successfully distill the complex discriminative logic from the massive Qwen3-VL-235B-A22B teacher model into the lightweight student evaluators, the design of the prompt is crucial. As discussed in the main text, we employ a Chain-of-Thought (CoT) strategy to prevent the teacher model from outputting arbitrary scalar values. 

The prompts are explicitly structured to force the model to generate a detailed, step-by-step semantic rationale (the \texttt{explanation} field) prior to assigning the final discrete Likert scores (the \texttt{scores} field). This autoregressive generation order ensures that the final scores are deeply grounded in logical deduction. During the two-stage LoRA fine-tuning, the student models are trained using these exact prompt templates to emulate both the reasoning process and the scoring distribution. During inference, the student models follow this same CoT structure to maximize evaluation accuracy and interpretability.

The comprehensive prompt templates for both image and video modalities are provided below.

\vspace{1em}
\noindent\textbf{B.1 Teacher Prompt for Image Editing Evaluation}

\begin{quote}
\small\ttfamily
You are an expert image editing evaluation model. You will evaluate the quality of an edited image compared to the original image.

**Input Information:**\\
- Original Prompt: \{original\_prompt\}\\
- Edited Prompt: \{edited\_prompt\}\\
- Original Image: [First image]\\
- Edited Image: [Second image]

**Evaluation Task:**\\
Evaluate the edited image across the following four dimensions. For each dimension, provide a score from 0 to 5: 0 (Completely fails), 1 (Poor quality with major issues), 2 (Below average), 3 (Acceptable quality), 4 (Good quality with very minor flaws), 5 (Excellent quality, meets all requirements).

**Evaluation Dimensions:**\\
1. **Structural Fidelity**: Evaluate whether the structure, pose, orientation, and spatial relationships of unedited entities remain perfectly consistent with the original image.\\
2. **Text-Image Alignment**: Evaluate how accurately the edited content reflects the requested changes described in the edited prompt.\\
3. **Background Consistency**: Evaluate the consistency of all regions EXCEPT the edited subject. Check for unwanted changes, color shifts, or distortions in unedited areas.\\
4. **Naturalness**: Evaluate whether the overall scene appears natural. Look for noticeable flaws such as inconsistent lighting, perspective errors, structural distortions, or watermarks.

**Important Guidelines:**\\
- [CRITICAL] You must FIRST provide a detailed step-by-step explanation for each dimension, and THEN output the final numerical scores.\\
- Be critical and use the full range of scores (0-5). A perfect score (5) should be rare and reserved for truly excellent results. Scores below 2 indicate significant failures.

**Output Format:**\\
Provide your evaluation strictly in the following JSON format:
\begin{verbatim}
{
  "explanation": {
    "structural_fidelity": "<reasoning>",
    "text_image_alignment": "<reasoning>",
    "background_consistency": "<reasoning>",
    "naturalness": "<reasoning>"
  },
  "scores": {
    "structural_fidelity": <score>,
    "text_image_alignment": <score>,
    "background_consistency": <score>,
    "naturalness": <score>
  }
}
\end{verbatim}
\end{quote}

\vspace{1em}
\noindent\textbf{B.2 Teacher Prompt for Video Editing Evaluation}

\begin{quote}
\small\ttfamily
You are an expert video editing evaluation model. You will evaluate the quality of an edited video compared to the original video.

**Input Information:**\\
- Original Prompt: \{original\_prompt\}\\
- Edited Prompt: \{edited\_prompt\}\\
- Original Video: [First video]\\
- Edited Video: [Second video]

**Evaluation Task:**\\
Evaluate the edited video across the following five dimensions. For each dimension, provide a score from 0 to 5: 0 (Completely fails), 1 (Poor quality with major issues), 2 (Below average), 3 (Acceptable quality), 4 (Good quality with very minor flaws), 5 (Excellent quality, meets all requirements).

**Evaluation Dimensions:**\\
1. **Structural Fidelity**: Evaluate whether the structure, pose, orientation, and spatial relationships of unedited entities remain consistent with the original video.\\
2. **Text-Video Alignment**: Evaluate how accurately the edited content reflects the requested changes described in the edited prompt.\\
3. **Background Consistency**: Evaluate the consistency of all regions EXCEPT the edited subject. Check for unwanted changes, color shifts, or distortions in unedited areas.\\
4. **Naturalness**: Evaluate whether the overall video appears natural, checking for flaws like inconsistent lighting, perspective errors, or structural distortions.\\
5. **Temporal-Spatial Consistency**: Focus on the continuity and logical rationality of the video across time and space. Check if the object maintains identity consistency, if motion trajectories are smooth, and if spatial logic (e.g., gravity, depth) is reasonable. Penalize any flickering, teleportation, or physical violations.

**Important Guidelines:**\\
- [CRITICAL] You must FIRST provide a detailed step-by-step explanation for each dimension, and THEN output the final numerical scores.\\
- Be critical and use the full range of scores (0-5). A perfect score (5) should be rare. Consider the intrinsic difficulty of the editing task while maintaining consistent scoring standards.

**Output Format:**\\
Provide your evaluation strictly in the following JSON format:

\begin{verbatim}
{
  "explanation": {
    "structural_fidelity": "<reasoning>",
    "text_video_alignment": "<reasoning>",
    "background_consistency": "<reasoning>",
    "naturalness": "<step-by-step reasoning>",
    "temporal_spatial_consistency": "<reasoning>"
  },
  "scores": {
    "structural_fidelity": <score>,
    "text_video_alignment": <score>,
    "background_consistency": <score>,
    "naturalness": <score>,
    "temporal_spatial_consistency": <score>
  }
}
\end{verbatim}

\end{quote}

\section{Qualitative Examples Across Task Taxonomy}
\label{appendix:visualizations}

To supplement the quantitative findings in the main manuscript, this appendix provides comprehensive qualitative visualizations across UniEditBench's full task taxonomy. These examples demonstrate the benchmark's unique capability to assess diverse, fine-grained editing operations, ranging from simple appearance-level attribute modifications to complex compositional and spatial reasoning tasks. 

By presenting results across multiple state-of-the-art models from distinct paradigms, we visualize the current performance bottlenecks in automated visual editing—especially concerning spatial relations and fine-grained text alignment—which often remain hidden when analyzing holistic scalar metrics.

\vspace{0.5em}
\noindent\textbf{C.1 Qualitative Image Editing Visualization:}
Figure A.1 displays generated results from representative models across all nine image editing categories defined in our visually-grounded taxonomy: Add, Adjust, Change, Count, Extract, Remove, Reorder, Replace, and Stroke. To showcase paradigm diversity, this visualization covers examples from Inversion-based, Inversion-free, Structured Editing, and MLLM-based methods. These snapshots illustrate distinct failure modes, such as counting errors, object merging during reordering, and background distortions, specific to each task and paradigm.

\vspace{0.5em}
\noindent\textbf{C.2 Qualitative Video Editing Visualization:}
Figure A.2 presents visualization results across the full video editing taxonomy, covering all eight defined categories: Add, Adjust, Change, Count, Extract, Remove, Reorder, Replace. These complex operations—especially compositional reasoning tasks like Count and Reorder which require both object permanence and logical logical motion flow—are visualized across three major paradigms including Inversion-free, Structured Editing, and MLLM-based video models. By presenting key frames, the visualization highlights the critical challenge of maintaining strict temporal-spatial consistency while executing precise semantic edits.

% --- Figure 1: Full Image Qualitative ---
\begin{figure*}[htbp]
  \centering
  \includegraphics[width=\textwidth]{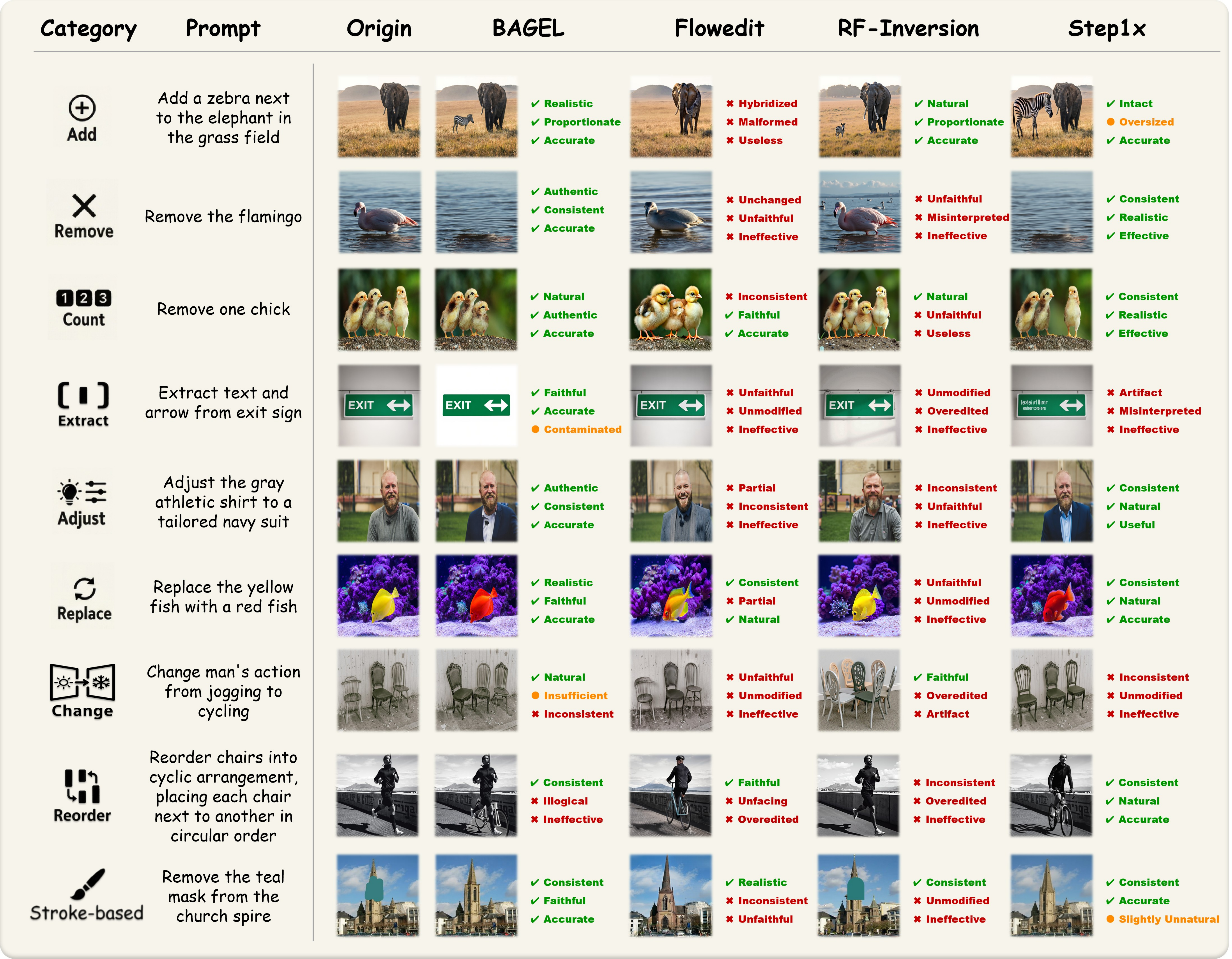}
  \caption{Comprehensive qualitative comparison on \textbf{Image Editing}. The examples cover the full 9-category taxonomy on the benchmark across four major paradigms: Inversion-based, Inversion-free, Structured Editing, and MLLM.}
  \label{fig:app_image_visual}
\end{figure*}

% --- Figure 2: Full Video Qualitative ---
\begin{figure*}[htbp]
  \centering
  \includegraphics[width=\textwidth]{2_appendix_2_baseline.jpg.pdf}
  \caption{Comprehensive qualitative comparison on \textbf{Video Editing}. The examples cover the full 8-category taxonomy on the benchmark across three major paradigms: Inversion-free, Structured Editing, and MLLM.}
  \label{fig:app_video_visual}
\end{figure*}

\section{Training Framework: ms-swift}
\label{appendix:msswift}

To implement the two-stage knowledge distillation described in Section \ref{sec:evaluator}, we utilized the ms-swift framework \cite{msswift}. Developed by the ModelScope community, ms-swift is a comprehensive training and deployment infrastructure that supports over 1,000 large language models and multi-modal large language models, including the Qwen3-VL architecture used in our study.

Specifically, we leveraged ms-swift to execute Parameter-Efficient Fine-Tuning via LoRA. The framework natively integrates advanced memory optimization and distributed training technologies, such as Flash-Attention, DeepSpeed ZeRO optimizations, and sequence parallelism. By utilizing these features, we efficiently distilled the 235B teacher model's reasoning logic and scoring distribution into the 4B and 8B student evaluators. This approach significantly reduced VRAM requirements, rendering high-fidelity multi-modal fine-tuning computationally feasible within standard academic hardware constraints.

\end{document}